\documentclass[10pt,twocolumn,letterpaper]{article}

\usepackage{iccv}
\usepackage{times}
\usepackage{epsfig}
\usepackage{graphicx}
\usepackage{amsmath}
\usepackage{amssymb}

\usepackage{url}
\usepackage{graphics}
\usepackage{color}
\usepackage{mathtools}
\usepackage{multirow}
\usepackage{subfigure}
\usepackage{tabularx}
\usepackage{wrapfig,lipsum,booktabs}
\usepackage{xcolor}
\usepackage{floatflt}
\usepackage{caption}

\usepackage[pagebackref=true,breaklinks=true,letterpaper=true,colorlinks,bookmarks=false,citecolor=cyan]{hyperref}
\usepackage[ruled, lined, linesnumbered, commentsnumbered, longend]{algorithm2e}
\newcommand{\vect}[1]{\mathbf{#1}}

\usepackage{caption}
\usepackage{enumitem}
\setitemize{noitemsep,topsep=0pt,parsep=0pt,partopsep=0pt}
\usepackage{flushend}

\iccvfinalcopy 

\def\iccvPaperID{2248} 

\ificcvfinal\fi

\begin{document}

\title{Transformer-Based Attention Networks for Continuous Pixel-Wise Prediction }

\author{Guanglei Yang$^{1,2}$ \quad Hao Tang$^3$ \quad Mingli Ding$^1$ \quad Nicu Sebe$^2$ \quad Elisa Ricci$^{2,4}$\\
$^1$Harbin Institute of Technology, China  \quad $^2$DISI, University of Trento, Italy \\
$^3$Computer Vision Lab, ETH Zurich, Switzerland  \quad $^4$Fondazione Bruno Kessler, Italy\\
{\tt\small \{yangguanglei,dingml\}@hit.edu.cn,\quad hao.tang@vision.ee.ethz.ch,\quad \{nicu.sebe,e.ricci\}@unitn.it}
}

\maketitle
\ificcvfinal\thispagestyle{empty}\fi

\begin{abstract}
While convolutional neural networks have shown a tremendous impact on various computer vision tasks, they generally demonstrate limitations in explicitly modeling long-range dependencies due to the intrinsic locality of the convolution operation. 
Initially designed for natural language processing tasks, Transformers have emerged as alternative architectures with innate global self-attention mechanisms to capture long-range dependencies.
In this paper, we propose TransDepth, an architecture that benefits from both convolutional neural networks and transformers.
To avoid the network losing its ability to capture local-level details due to the adoption of transformers, we propose a novel decoder that employs attention mechanisms based on gates. Notably, this is the first paper that applies transformers to pixel-wise prediction problems involving continuous labels (\textit{i.e.}, monocular depth prediction and surface normal estimation).
Extensive experiments demonstrate that the proposed TransDepth achieves state-of-the-art performance on three challenging datasets. Our code is available at: \url{https://github.com/ygjwd12345/TransDepth}.
\end{abstract}
\section{Introduction}
Over the past decade, convolutional neural networks have become the privileged methodology to address fundamental and challenging computer vision tasks requiring dense pixel-wise prediction, such as semantic segmentation~\cite{chen2016attention,fu2019dual}, monocular depth prediction ~\cite{liu2015deep,eigen2014depth}, and normal surface computation \cite{qi2018geonet}.
Since the seminal work of \cite{he2016deep},
existing depth prediction models' have been dominated by encoders implemented with architectures such as ResNet and VGG-Net. The encoder progressively reduces the spatial resolution and learns more concepts with larger receptive fields. Because context modeling is critical for pixel-level prediction, deep feature representation learning is arguably the most critical model component~\cite{chen2017deeplab}. However, it is still challenging for depth prediction networks to improve their ability in modeling global contexts. Traditionally, both stacked convolution layers and consecutive down-sampling are used in the encoders to generate sufficiently large receptive fields of deep layers. This problem is typically circumvented rather than resolved to some extent. Unfortunately, existing strategies bring several drawbacks: (1) the training of very deep nets is affected by the fact that consecutive multiplications wash out low-level features; (2) the local information crucial to dense prediction tasks is discarded since the spatial resolution is reduced gradually. To overcome these limitations, several methods have been recently proposed. One solution is manipulating the convolutional operation directly by using for example large kernel sizes~\cite{peng2017large}, atrous convolutions~\cite{chen2017deeplab}, and image/feature pyramids~\cite{zhao2017pyramid}. Another solution is to integrate attention modules into the fully convolutional network architecture. Such a module aims to model global interactions of all pixels in the feature map~\cite{wang2018non}. When applied to monocular depth prediction~\cite{xu2018pad,xu2020probabilistic} a general approach is to combine the attention module with a multi-scale fusion method.
More recently, Huynh~\etal~\cite{huynh2020guiding} proposed a depth-attention volume to incorporate a non-local coplanarity constraint to the network.
Guizilini~\etal~\cite{guizilini2020semantically} rely on a fixed pre-trained semantic segmentation network to guide global representation learning. Though these methods' performance is improved significantly, still the above mentioned issues persist. 

Transformers were initially used to model sequence-to-sequence predictions in NLP tasks to obtain a larger receptive field and have recently attracted tremendous interest in the computer vision community. The first purely self-attention-based Vision Transformer (ViT) for image recognition was proposed in~\cite{dosovitskiy2020image} attaining excellent results on ImageNet compared with the convolutional networks. 
Moreover, SETR~\cite{zheng2020rethinking} replaces the encoders with pure Transformers, obtaining competitive results on the CityScapes dataset. Interestingly, we found that a SETR-like pure Transformer-based segmentation network produces unsatisfactory performance due to the lack of spatial inductive bias in modeling the local information. Meanwhile, most previous methods based on deep feature representation learning fail to solve this problem. 
Nowadays, only few researchers~\cite{carion2020end} are considering combining the CNNs with Transformers to create a hybrid structure to combine their advantages.

In contrast to treating pixel-level prediction tasks as a sequence-to-sequence prediction problem, we firstly propose to embed Transformers into the ResNet backbone in order to model semantic pixel dependencies. Moreover, we design a new and effective unified attention gate decoder to address the drawback that the pure linear Transformer's embedding feature lacks spatial inductive bias in capturing the local representation. We show empirically that our method offers a new perspective in model design and achieves state-of-the-art on several challenging benchmarks.

To summarize, our contribution is threefold:
\begin{itemize}[leftmargin=*]
\item We are the first to propose the use of Transformers for both monocular depth estimation and surface normal prediction tasks.  Transformers can successfully improve the ability of traditional convolutional neural networks to model long-range dependencies.
\item We propose a novel and effective unified attention gate structure designed to utilize and fuse multi-scale information in a parallel manner and pass information among different affinities maps in the attention gate decoders for better modeling the multi-scale affinities.
\item We conduct extensive experiments on two distinct pixel-wise prediction tasks with three challenging datasets (\eg, NYU~\cite{silberman2012indoor}, KITTI~\cite{Geiger2013IJRR}, and ScanNet~\cite{dai2017scannet}), demonstrating that our TransDepth outperforms previous methods on KITTI (0.956 on $\delta \textless 1.25$), NYU depth (0.900 on $\delta \textless 1.25$), and achieves new state-of-the-art results on NYU surface normal estimation.
\end{itemize}
\section{Related Work}

\noindent\textbf{Transformers in Computer Vision.}
Transformer and self-attention models have revolutionized machine translation and natural language processing~\cite{vaswani2017attention,dai2019transformer}.
Recently, there were also some explorations for the usage of Transformer structures in computer vision tasks~\cite{hu2019local,carion2020end,liu2021end,ding2021looking,yang2021transformer,ren2021cloth}. 
For instance, 
LRNet~\cite{hu2019local} explored local self-attention to avoid the heavy computation brought by global self-attention. 
Axial-Attention~\cite{wang2020axial} decomposed the global spatial attention into two separate axial attention such that the computation is vastly reduced. Apart from these pure Transformer-based models, there are also CNN-Transformer hybrid ones.
For instance, DETR~\cite{carion2020end} and the following deformable version utilized a Transformer for object detection where the Transformer was appended inside the detection head. 
LSTR~\cite{liu2021end} adopted Transformers for disparity estimation and for lane shape prediction. 
Most recently, ViT~\cite{dosovitskiy2020image} was the first work to show that a pure Transformer-based image classification model can achieve the state-of-the-art. This work provides a direct inspiration to exploit a pure Transformer-based encoder design in a semantic segmentation model. Meanwhile, SETR  \cite{zheng2020rethinking} based on ViT, leverages attention for image segmentation. However, there is no related work in continuous pixel prediction. The main reason is that the networks, designed for the continuous label task, extremely rely on deep representation learning and fully-convolutional networks (FCN) with a decoder architecture.
In this case, the pure Transformer (without convolution and resolution reduction) regarding an image as a patch sequence is unsuitable for pixel-level prediction with continual labels.

We propose a novel combination framework to put a linear Transformer and ResNet together to address the limitation mentioned above.
It leads to that the previous effective methods based on deep representing learning, such as dilated/atrous convolutions and inserting attention modules, are still compatible with our networks. Meanwhile, the position embedding module is removed from our linear Transformer, but we take advantage of multi-scale fusion in the decoder to add position information.  It is essential to successfully apply Transformers to depth prediction and surface normal estimation tasks.

\noindent\textbf{Monocular Depth Estimation.} 
Most recent works on monocular depth estimation are based on CNNs~\cite{eigen2015predicting,liu2015deep,wang2015towards,laina2016deeper, fu2018deep,lee2019big,guizilini20203d,guizilini2020semantically,xu2018structured}, which suffer from the limited receptive field problem or from the less global representation learning. For instance,
Eigen~\etal~\cite{eigen2014depth} introduced a two-stream deep network to take into account both coarse global prediction and local information. 
Fu~\etal~\cite{fu2018deep} proposed a discretization strategy to treat monocular depth estimation as a deep ordinal regression problem. They also employed a multi-scale network to capture relevant multi-scale information.
Lee~\etal~\cite{lee2019big} introduced local planar
guidance layers in the network decoder module to learn more effective features for depth estimation.
More recently, PackNet-SfM~\cite{guizilini20203d} used 3D convolutions with self-supervision to learn detail-preserving representations. At the same time, Guizilini~\etal~\cite{guizilini2020semantically} exploit semantic features into the self-supervised depth network by using a pre-trained semantic segmentation network. The new SOTA, FAL-Net~\cite{gonzalez2020forget}, focuses instead on representation learning using stereoscopic view synthesis penalizing the synthetic right-view in all image regions. Though it explicitly increases long-range modeling dependencies, more training steps are added.

\begin{figure*}[htb] \small
\centering
\includegraphics[width=0.9\textwidth]{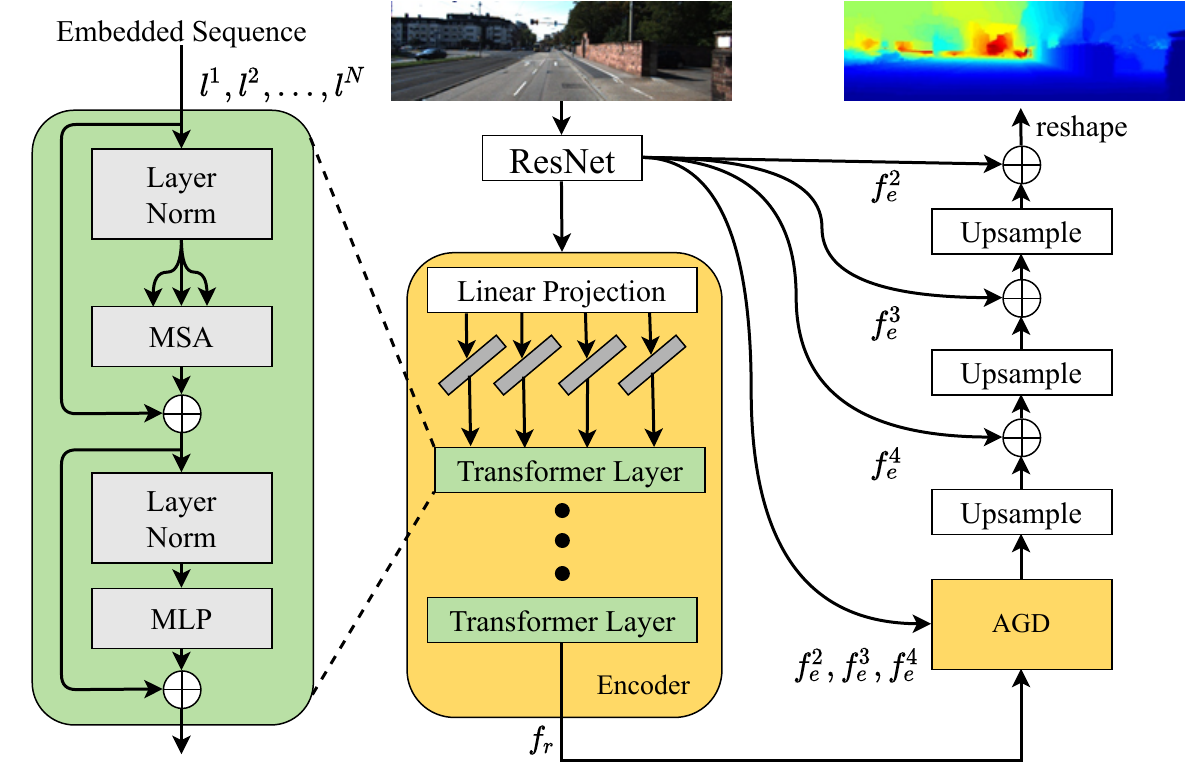}
\caption{The overview of the proposed TransDepth. The symbols \textcircled{c} and $\oplus$ denote concatenation and addition operations, respectively. AG is short for attention gate.}
\label{fig:overview}
\vspace{-0.4cm}
\end{figure*}

Our method focuses on representation learning as well but with only one step training strategy. The Transformer mechanism is quite suitable to solve the limited receptive field issue, to guide the generation of depth features. Unlike the previous works~\cite{zheng2020rethinking, dosovitskiy2020image} reshaping the image into a sequence of flattened 2D patches, we propose a hybrid model combining ResNet~\cite{he2016deep} and linear Transformer~\cite{dosovitskiy2020image}. 
This is quite different from the previous Transformer mechanism, taking advantage of both sides. This composite structure also holds another advantage: many deep representation learning methods can be easily transferred in this network.

\noindent\textbf{Surface Normal Estimation.}
Surface normal prediction is regarded as a close task to monocular depth prediction.
Extracting 3D geometry from a single image has been a longstanding problem in computer vision. Surface normal estimation is a classical task in this context requiring modeling both global and local features. Typical approaches leverage networks with high capacity to achieve accurate predictions at high resolution. For instance, FrameNet \cite{huang2019framenet} employed the DORN \cite{fu2018deep} architecture, a modification of DeepLabv3 \cite{chen2017deeplab} that removes multiple spatial reductions (2$\times$2 max pool layers), to generate high resolution surface normal maps. A different strategy consists of designing appropriate loss terms. For instance, UprightNet \cite{xian2019uprightnet} considered an angular loss and showed its effectiveness for the task. More recently, Do~\etal~\cite{do2020surface} proposed a novel truncated angular loss and a tilted image process, keeping the atrous spatial pyramid pooling (ASPP) module to increase the receptive field. Although its performance is SOTA, two extra training phases are added due to the tilted image process.

\noindent\textbf{Attention Models.}
Several works have considered integrating attention models within deep architectures to improve performance in several tasks, such as image categorization~\cite{xiao2015application}, image generation \cite{tang2020xinggan,tang2020dual,tang2019multi,tang2019attention}, video generation~\cite{liu2021cross}, speech recognition~\cite{chorowski2015attention}, and machine translation~\cite{vaswani2017attention}. Focusing on pixel-level prediction, Chen \etal~ \cite{chen2016attention} were the first to describe an attention model to combine multi-scale features learned by a FCN for semantic segmentation. 
Zhang \etal~\cite{zhang2018context} designed EncNet, a network equipped with a channel
attention mechanism to model the global context. 
Huang \etal~\cite{huang2019ccnet} described CCNet, a deep architecture that embeds a criss-cross attention module with the idea of modeling contextual dependencies using sparsely connected graphs to achieve higher computational efficiency. 
Fu \etal~\cite{fu2019dual} proposed to model semantic dependencies associated with spatial and channel dimensions by using two separate attention modules. 

Our work significantly departs from these approaches as we introduce a novel attention gate mechanism, adding spatial- and channel-level attention into the attention decoder. Notably, we also prove that our model can be successfully employed in the case of several challenging dense continual pixel-level prediction tasks, where it significantly outperforms PGA-Net~\cite{xu2020probabilistic}.
\section{The Proposed TransDepth}

\begin{figure*}[t] \small
\centering
\includegraphics[width=1\textwidth]{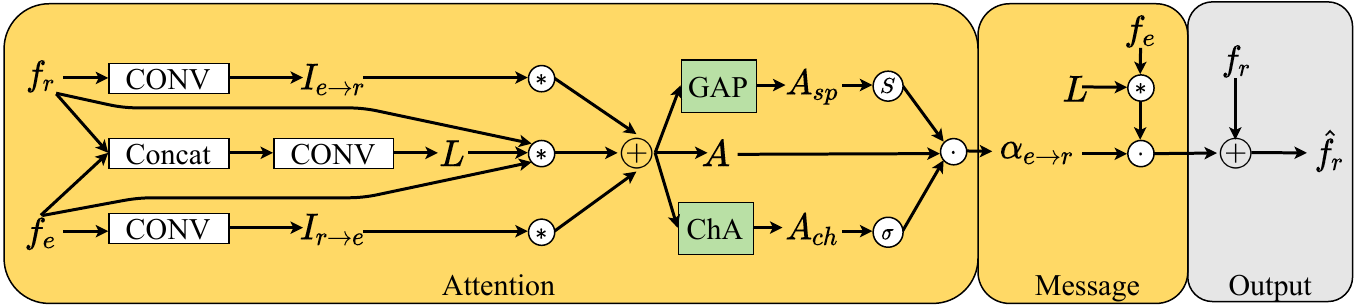}
\caption{The overview of the proposed attention gate module. The symbols $\odot$, $\oplus$, \textcircled{$\sigma$}, \textcircled{$\ast$}, and \textcircled{\tiny{S}} denote element-wise multiplication, element-wise addition, sigmoid, convolution, and softmax operation, respectively.
}
\label{fig:agdeconder}
\vspace{-0.4cm}
\end{figure*}

As previously discussed, our work aims to solve limited receptive fields by adding Transformer layers and enhancing the learned representation by an attention gate decoder. 
\subsection{Transformer for Depth Prediction}

An overview of the network is depicted in Figure~\ref{fig:overview}. Unlike the previous works~\cite{zheng2020rethinking,chen2021transunet, dosovitskiy2020image} reshaping the image $I{\in} \mathbb{R}^{H\times W \times 3}$ into a sequence of flattened 2D patches $I_p{\in} \mathbb{R}^{N\times (p^2\cdot 3)}$, we propose a hybrid model. As shown in Figure~\ref{fig:overview}, the input sequence comes from a ResNet backbone~\cite{he2016deep}. 
Then the patch embedding is applied to patches extracted from the final feature output of a CNN. This patch embedding's kernel size should be $p{\times} p$, which means that the input sequence is obtained by simply flattening the spatial dimensions of the feature map and projecting to the Transformer dimension. In this case, we also remove position embedding because the original physical meaning is missing while mapping the vectorized patches $I_p$ into a latent embedding space $l_p$ using a linear projection. The input of the first Transformer layer is calculated as follow:
\begin{equation}
    z_0=[l^1E; l^2E; \cdots; l^NE],
\end{equation}
where $z_0$ is mapped into a latent N-dimensional embedding space using a trainable linear projection layer and $E$ is the patch embedding projection. There are $L$ Transformer layers which consist of multi-headed self-attention (MSA) and multi-layer perceptron (MLP) blocks. At each layer $\ell$, the input of the self-attention block is a triplet of $Q$ (query), $K$ (key), and $V$ (value), similar with~\cite{vaswani2017attention}, computed from $z_{\ell-1}{\in} \mathbb{R}^{L\times C}$ as:
\begin{equation}
    Q= z_{\ell-1}\times W_Q, K= z_{\ell-1}\times W_K , V=z_{\ell-1}\times W_V,
\end{equation}
where $W_Q,W_K,W_V{\in}\mathbb{R}^{C\times d}$ are the learnable parameters of weight matrices and $d$ is the dimension of $Q$, $K$, $V$. The self-attention is calculated as:
\begin{equation}
    {\rm AH}= {\rm softmax}(\frac{Q\times K^T}{\sqrt{d}})\cdot V, 
\end{equation}
where AH is short for attention head and $d$ is the dimension of self-attention block.
MSA means the attention head will be calculated $m$ times by independent weight matrices. The final ${\rm MSA}(z_{\ell-1})$ is defined as:
\begin{equation}
    {\rm MSA}(z_{\ell-1}) = z_{\ell-1} + {\rm concat}({\rm AH}_1; {\rm AH}_2; \cdots; {\rm AH}_m)\times W_o,
\end{equation}
where $W_o{\in} \mathbb{R}^{md\times C}$. 
The output of ${\rm MSA}$ is then transformed by a MLP block with residual skip as the layer output as:
\begin{equation}
    z_{\ell}= {\rm MLP}({\rm LN}(z'_{\ell})) + z'_{\ell},
\end{equation}
where ${\rm LN}(\cdot)$ means the layer normalization operator and $z'_{\ell}{=} {\rm MSA}(z_{\ell-1})$. The structure of a Transformer layer is illustrated in the left part of Figure~\ref{fig:overview}. After the Transformer layer, the output will be recovered to the original feature shape. 

\subsection{Attention Gate Decoder}
Given an input image $\vect{I}$ and a generic front-end CNN model, we consider a set of 
$S$ multi-scale feature maps $\vect{F} {=} \{f^{i}\}_{i=1}^{N}$. Being a generic framework, these feature maps can 
be the output of $S$ intermediate CNN layers or of another representation, thus $s$ is a \textit{virtual} scale. 
Opposite to previous works adopting simple 
concatenation or weighted averaging schemes 
\cite{zheng2020rethinking}, we propose to combine the multi-scale feature maps by learning a 
set of latent kernels ($I_{r\to e}$, $I_{e\to r}$, $L$) with a novel structure Attention-Gated 
module sketched in Figure~\ref{fig:agdeconder}. We choose $f^N$ as a receive feature only, $f_r$, while $\{f^{i}\}_{i=1}^{N-1}$ are chosen as emitting features, $f_e$, in all tasks. The influence of the fusion of different scales is explained in the ablation part. 

In detail, the whole attention gate can be divided into two parts, \ie, attention and message. We propose to bring together recent advances in pixel-wise prediction by formulating a novel attention gate mechanism for the attention part. Inspired by~\cite{fu2019dual}, where two spatial- and channel-wise predictions are computed, we opt to infer different spatial and channel attention variables.
Our attention tensor can be defined by:
\begin{equation}
\label{eq:structured-attention}
\begin{aligned}
 A_{sp}^i &= \frac{1}{C}\sum_{c=1}^{C}(\omega_{sp}*A^i)[c,h,w], \\
 A_{ch}^i &= \frac{1}{HW}\sum_{h,w=1}^{H,W}(\omega_{sp}*A^i)[c,h,w],\\
 \alpha_{e\to r}^i &= {\rm softmax}(A_{sp}^i) \cdot \sigma(A_{ch}^i) \cdot A^i,
\end{aligned}
\end{equation}
where $i$ means $f^i$ is chosen as an emitting feature.
Different from~\cite{fu2019dual}, we adapt a local conditional kernel before generating attention. The kernels $I_{r\to e}$, $I_{e\to r}$, and $L$ are
predicted from the input features using a linear transformation as follows:
\begin{equation}\label{eq:featureconditionalkernel}
\begin{aligned}
\vect{L}^{i,j} &=  {\vect{W}_L}^{i,j} \mathrm{concat}(f_e^i, f_{r}^j) + {\vect{b}_L}^{i,j},\\
\vect{I}_{r\to e}^{i,j} &=  {\vect{W}_I}_{r\to e}^{i,j}\ f_e^i + {\vect{b}_I}_{r\to e}^{i,j}, \\
\vect{I}_{e\to r}^{i,j} &= {\vect{W}_I}_{e\to r}^{i,j} f_{r}^j + {\vect{b}_I}_{e\to r}^{i,j}.
\end{aligned}
\end{equation}
Then, the integrated attention is defined as follow:
\begin{equation}
    A^i = \vect{I}_{e\to r}^i*f_r+\vect{I}_{r\to e}^i*f_e^i + f_r*L*f_e^i.
\label{eq:full attention}
\end{equation}
Compared with the attention part, the message is easy to be calculated by $L^i*f_r$. Finally, the output of our attention gate decoder is:
\begin{equation}
    \hat{f}_e^i={\rm concat}(L^1*f_e^1\cdot \alpha_{e\to r}^1+f_r,...,L^{N-1}*f_e^{N-1}\cdot \alpha_{e\to r}^{N-1}+f_r).
\label{eq:full attention}
\end{equation}

Once the hidden variables are updated, we use them to address several different discrete prediction tasks, including monocular depth estimation and surface normal estimation. Following previous works, the network optimization loss for depth prediction, updated from~\cite{eigen2014depth}, is:
\begin{equation}
    \mathcal{L}_{depth}=\alpha\sqrt{\frac{1}{T}\sum_i g_i^2-\frac{\lambda}{T^2}(\sum_i g_i)^2},
\label{eq:depth loss}
\end{equation}
where $g_i {=} \log\hat{d}_i{-}\log d_i$ with the ground truth depth $d_i$ and the predicted depth $\hat{d}_i$.
We set $\lambda$ and $\alpha$ to 0.85 and 10,
same with~\cite{lee2019big}. The angular loss is chosen as the surface normal loss. 
\section{Experiments}
\subsection{Datasets}
The NYU dataset~\cite{silberman2012indoor} is used to evaluate our approach in the depth estimation task. We use 120K RGB-Depth pairs with a resolution of $480{\times}640$ pixels, acquired with a Microsoft Kinect device from 464 indoor scenes. We follow the standard train/test split as in~\cite{eigen2014depth}, using 249 scenes for training and 215 scenes (654 images) for testing. We also use this dataset to evaluate our approach in the surface normal task, including 795 training images and 654 testing images.

The KITTI dataset~\cite{Geiger2013IJRR} is a large-scale outdoor dataset created for various autonomous driving tasks. We use it to evaluate the depth estimation performance of our proposed model. Following the standard training/testing split proposed by Eigen~\etal~\cite{eigen2014depth}, we specifically use 22,600 frames from 32 scenes for training and 697 frames from the rest 29 scenes for testing.

The ScanNet dataset~\cite{dai2017scannet} is a large RGB-D dataset for 3D scene understanding. We employ it to evaluate the surface normal performance of our proposed model. ScanNet dataset is divided into 189,916 for training and 20,942 for testing with file lists provided in \cite{dai2017scannet}.

\subsection{Evaluation Metrics}
\par\noindent\textbf{Evaluation Protocol on Monocular Depth Estimation.}
We follow the standard evaluation protocol as in previous works~\cite{eigen2015predicting,eigen2014depth,wang2015towards} and adopt the following quantitative evaluation metrics in our experiments: 
\begin{itemize}[leftmargin=*]
\item Abs relative error (abs-rel): 
\( \frac{1}{K}\sum_{i=1}^K\frac{|\tilde{d}_i - d_i^\star|}{d_i^\star} \); 
\item Squared Relative difference (sq-rel):
\( \frac{1}{K}\sum_{i=1}^K\frac{||\tilde{d}_i - d_i^\star||^2}{d_i^\star} \); 
\item Root mean squared error (rms): 
\( \sqrt{\frac{1}{K}\sum_{i=1}^K(\tilde{d}_i - d_i^\star)^2} \);
\item Mean log10 error (log-rms): 
\( \sqrt{\frac{1}{K}\sum_{i=1}^K \Vert \log_{10}(\tilde{d}_i) - \log_{10}(d_i^\star) \Vert^2 }\);
\item Accuracy with threshold $t$: percentage (\%) of $d_i^\star$, subject to $\max (\frac{d_i^\star}{\tilde{d}_i}, \frac{\tilde{d}_i}{d_i^\star}) {=} 
\delta {<} t~(t {\in} [1.25, 1.25^2, 1.25^3])$;
\end{itemize}
where $\tilde{d}_i$ and $d_i^\star$ is the ground-truth depth and the estimated depth at pixel $i$ respectively; $K$ is the total number of pixels of the test images.

\par\noindent\textbf{Evaluation Protocol on Surface Normal Estimation.} We utilize five standard evaluation metrics~\cite{fouhey2013data}. For space limitation, we pick up median angle distance between prediction and ground-truth for valid pixels and the fraction of pixels with angle difference with ground-truth less than $11.25^{\circ}$ listed in the main paper. The results of five standard evaluation metrics are put into supplementary material.

\subsection{Implementation Details} 
The proposed TransDepth is implemented in~PyTorch. The experiments are conducted on four Nvidia Tesla V100 GPUs, each with 32 GB memory. The ResNet-50 architecture pretrained on ImageNet~\cite{deng2009imagenet} is considered in the experiments for initializing the backbone network of our encoder network. For parameters of Transformer, T-layers, Hidden size, and attention multi-head are set to 12, 768, and 12, respectively.
As the attention gate decoder structure setting, $f^5$ is chosen as the receiving feature, $f_r$, while $\{f^{i}\}_{i=3}^{5}$ are taken up as emitting features, $f_e$, in all tasks.

For the monocular depth estimation and surface normal prediction tasks, the learning rate is set to $10^{-4}$ with a weight decay of 0.01. The Adam optimizer is used in all our experiments with a batch size of 16 for all tasks. 
The total training epochs are set to 50 for depth prediction and 20 for surface normal prediction. We train our network on a random crop of size 352$\times$704 for KITTI dataset, 416$\times$512 for NYU dataset for depth prediction while
the input image size is uniformly set to 320$\times$256 for surface normal prediction.

\begin{figure*}[!h] \small
\centering
\subfigure[Image]{
    \begin{minipage}{0.14\linewidth}
        \centering
        \includegraphics[width=0.993\textwidth,height=0.6in]{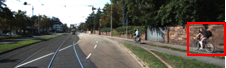}\\
        \includegraphics[width=0.993\textwidth,height=0.6in]{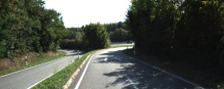}\\
        \includegraphics[width=0.993\textwidth,height=0.6in]{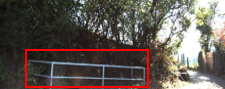}\\
        \includegraphics[width=0.993\textwidth,height=0.6in]{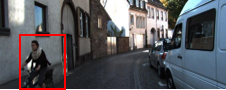}\\
        \includegraphics[width=0.993\textwidth,height=0.6in]{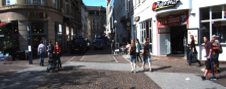}\\

    \end{minipage}%
}%
\subfigure[GT]{
    \begin{minipage}{0.14\linewidth}
        \centering
        \includegraphics[width=0.993\textwidth,height=0.6in]{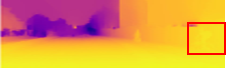}\\
        \includegraphics[width=0.993\textwidth,height=0.6in]{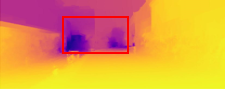}\\
        \includegraphics[width=0.993\textwidth,height=0.6in]{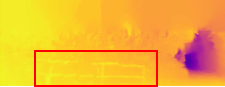}\\    
        \includegraphics[width=0.993\textwidth,height=0.6in]{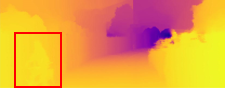}\\   
        \includegraphics[width=0.993\textwidth,height=0.6in]{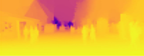}\\ 
    \end{minipage}%
}%
\subfigure[FAL-Net]{
    \begin{minipage}{0.14\linewidth}
        \centering
        \includegraphics[width=0.993\textwidth,height=0.6in]{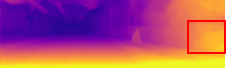}\\
        \includegraphics[width=0.993\textwidth,height=0.6in]{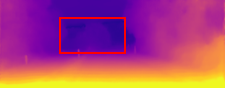}\\
        \includegraphics[width=0.993\textwidth,height=0.6in]{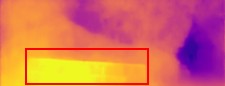}\\   
        \includegraphics[width=0.993\textwidth,height=0.6in]{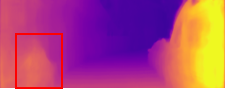}\\  
        \includegraphics[width=0.993\textwidth,height=0.6in]{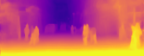}\\  

    \end{minipage}%
}%
\subfigure[Baseline]{
    \begin{minipage}{0.14\linewidth}
        \centering
        \includegraphics[width=0.993\textwidth,height=0.6in]{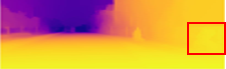}\\
        \includegraphics[width=0.993\textwidth,height=0.6in]{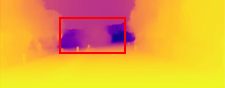}\\
        \includegraphics[width=0.993\textwidth,height=0.6in]{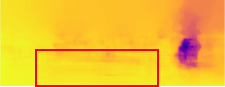}\\   
        \includegraphics[width=0.993\textwidth,height=0.6in]{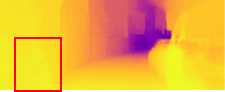}\\  
        \includegraphics[width=0.993\textwidth,height=0.6in]{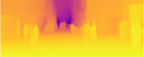}\\  

    \end{minipage}%
}%
\subfigure[Baseline w/ AGD]{
    \begin{minipage}{0.14\linewidth}
        \centering
        \includegraphics[width=0.993\textwidth,height=0.6in]{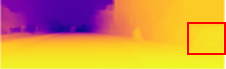}\\
        \includegraphics[width=0.993\textwidth,height=0.6in]{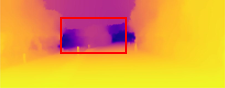}\\
        \includegraphics[width=0.993\textwidth,height=0.6in]{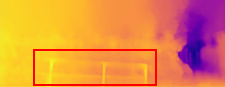}\\
        \includegraphics[width=0.993\textwidth,height=0.6in]{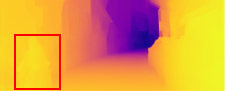}\\
        \includegraphics[width=0.993\textwidth,height=0.6in]{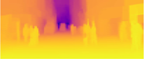}\\

    \end{minipage}%
}%
\subfigure[Baseline w/ ViT]{
    \begin{minipage}{0.14\linewidth}
        \centering
        \includegraphics[width=0.993\textwidth,height=0.6in]{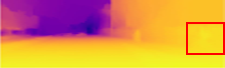}\\
        \includegraphics[width=0.993\textwidth,height=0.6in]{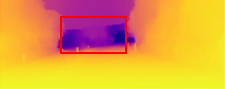}\\
        \includegraphics[width=0.993\textwidth,height=0.6in]{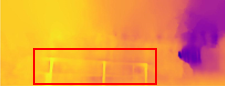}\\
        \includegraphics[width=0.993\textwidth,height=0.6in]{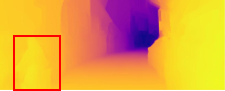}\\
        \includegraphics[width=0.993\textwidth,height=0.6in]{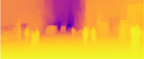}\\

    \end{minipage}%
}%
\subfigure[Ours (Full)]{
    \begin{minipage}{0.14\linewidth}
        \centering
        \includegraphics[width=0.993\textwidth,height=0.6in]{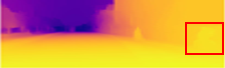}\\
        \includegraphics[width=0.993\textwidth,height=0.6in]{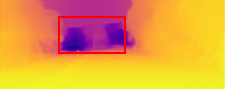}\\
        \includegraphics[width=0.993\textwidth,height=0.6in]{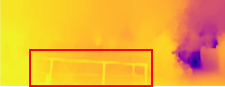}\\
        \includegraphics[width=0.993\textwidth,height=0.6in]{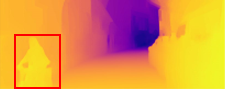}\\
        \includegraphics[width=0.993\textwidth,height=0.6in]{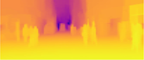}\\
    \end{minipage}%
}%
\centering
\caption{Qualitative examples on the KITTI dataset.}
\label{fig:vis_kitti}
\vspace{-0.4cm}
\end{figure*}

\begin{table*}[t] \small
\centering
\caption{Depth Estimation: KITTI dataset. K: KITTI. CS: CityScapes~\cite{cordts2016cityscapes}. CS$\to$K: CS pre-training. D: Depth supervision. M, Se, V, S: Monocular, segmentation, video, stereo. Sup: supervise.}
\label{tab:depth_kitti}
\begin{tabular}{rccccccccc}
\toprule[1.2pt]
\multirow{2}{*}{Method}& \multirow{2}{*}{Sup} & \multirow{2}{*}{Data} & \multicolumn{4}{c}{Error (lower is better)} & \multicolumn{3}{c}{Accuracy (higher is better)} \\ \cmidrule(lr){4-7} \cmidrule(lr){8-10} 
 & &  & abs rel & sq rel & rms & log rms & $\delta \textless 1.25$ & $\delta \textless 1.25^2$ & $\delta \textless 1.25^3$ \\
\midrule
CC~\cite{ranjan2019competitive}& M+Se & K & 0.140 & 1.070 & 5.326 & 0.217 & 0.826 & 0.941 & 0.975 \\
Bian~\etal \cite{bian2019unsupervised}& M+V & K+CS & 0.137 & 1.089 & 5.439 & 0.217 & 0.830 & 0.942 & 0.975 \\
DeFeat~\cite{spencer2020defeat}& M & K & 0.126 & 0.925 & 5.035 & 0.200 & 0.862 & 0.954 & 0.980 \\
$S^3$Net~\cite{cheng2020s}& M+Se & K & 0.124 & 0.826 & 4.981 & 0.200 & 0.846 & 0.955 & 0.982 \\
Monodepth2~\cite{godard2019digging}& M & K & 0.115 & 0.903 & 4.863 & 0.193 & 0.877 & 0.959 & 0.981 \\
pRGBD~\cite{tiwari2020pseudo}& M & K & 0.113 & 0.793 & 4.655 & 0.188 & 0.874 & 0.960 & 0.983 \\
Johnston~\etal \cite{johnston2020self}& M & K & 0.106 & 0.861 & 4.699 & 0.185 & 0.889 & 0.962 & 0.982 \\
SGDepth~\cite{klingner2020self}& M+Se & K+CS & 0.107 & 0.768 & 4.468 & 0.180 & 0.891 & 0.963 & 0.982 \\
Shu~\etal \cite{shu2020feature}& M & K & 0.104 & 0.729 & 4.481 & 0.179 & 0.893 & 0.965 & 0.984 \\
DORN~\cite{fu2018deep}& D & K & 0.072 & 0.307 & 2.727 & 0.120 & 0.932 & 0.984 & 0.994 \\
Yin~\etal \cite{yin2019enforcing}& M & K & 0.072 & - & 3.258 & 0.117 & 0.938 & 0.990 & 0.998 \\
PackNet~\cite{guizilini20203d}& V & K+CS & 0.071 & 0.359 & 3.153 & 0.109 & 0.944 & 0.990 & 0.997 \\
FAL-Net~\cite{gonzalez2020forget}& S & K+CS & 0.068 & 0.276 & 2.906 & 0.106 & 0.944 & 0.991 & 0.998 \\
PGA-Net~\cite{xu2020probabilistic} & D & K & 0.063 & 0.267 & \textbf{2.634} & 0.101 & 0.952 & 0.992 & 0.998\\ 
BTS \cite{lee2019big}& M & K & \textbf{0.061} & 0.261 & 2.834 & 0.099 & 0.954 & 0.992 & 0.998 \\
 \midrule
Baseline & M & K & 0.106 & 0.753 &  3.981 & 0.104 & 0.888  & 0.967  & 0.986\\
Ours w/ AGD & M & K & 0.065 & 0.261 & 2.766 & 0.101 & 0.953 & 0.993 & 0.998 \\
Ours w/ ViT & M & K & 0.064 & 0.258 & 2.761 & 0.099 & 0.955 & 0.993 & 0.999\\
Ours w/ AGD+ViT (Full) & M & K & 0.064 & \textbf{0.252} & 2.755 & \textbf{0.098} & \textbf{0.956} & \textbf{0.994} & \textbf{0.999} \\
\bottomrule[1.2pt]
\end{tabular}%
\vspace{-0.4cm}
\end{table*}

\subsection{Results on Monocular Depth Estimation}
We compare the proposed method with the leading monocular depth estimation models, \ie, \cite{ranjan2019competitive,bian2019unsupervised,spencer2020defeat,cheng2020s,godard2019digging,tiwari2020pseudo,johnston2020self,klingner2020self,shu2020feature,guizilini20203d,fu2018deep,yin2019enforcing,gonzalez2020forget,lee2019big}. 
Comparison results on the KITTI dataset are shown in Table~\ref{tab:depth_kitti}. 
Our method performs favorably versus all previous fully- and self-supervised methods, achieving the best results on the majority of the metrics. 
Our approach employs the supervised setting using single monocular images in the training and testing phase.
Compared with recent SOTA, \ie, FAL-Net, BTS, and PGA-Net, our method is better by a large margin. Meanwhile, unlike FAL-Net using stereo split, two-step training, and post-processing, our method is end-to-end without extra post-processing. The more important thing is that ``Ours w/ ViT'' has outperformed the SOTA.
It can support our standpoint that adding a linear Transformer makes networks improve their ability to capture long-range dependencies. In other words, our network becomes more straightforward but more potent by adapting the linear Transformer.

\begin{figure}[t] \small
\centering
\subfigure[Image]{
    \begin{minipage}{0.19\linewidth}
        \centering
        \includegraphics[width=0.993\textwidth,height=0.6in]{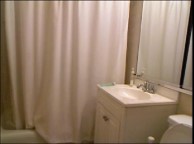}\\ 
        \includegraphics[width=0.993\textwidth,height=0.6in]{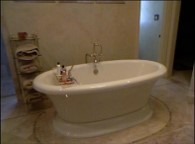}\\         \includegraphics[width=0.993\textwidth,height=0.6in]{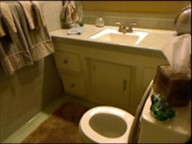}\\         \includegraphics[width=0.993\textwidth,height=0.6in]{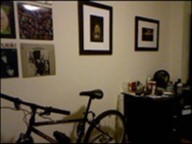}\\         \includegraphics[width=0.993\textwidth,height=0.6in]{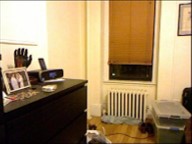}\\ 
    \end{minipage}%
}%
\subfigure[GT]{
    \begin{minipage}{0.19\linewidth}
        \centering

        \includegraphics[width=0.993\textwidth,height=0.6in]{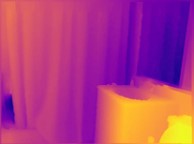}\\  
        \includegraphics[width=0.993\textwidth,height=0.6in]{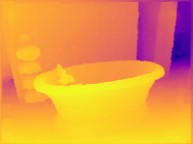}\\
        \includegraphics[width=0.993\textwidth,height=0.6in]{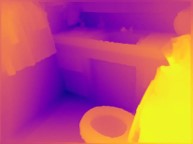}\\  
        \includegraphics[width=0.993\textwidth,height=0.6in]{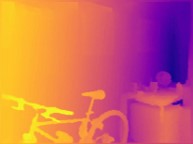}\\
        \includegraphics[width=0.993\textwidth,height=0.6in]{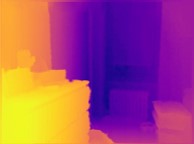}\\ 
    \end{minipage}%
}%
\subfigure[DORN]{
    \begin{minipage}{0.19\linewidth}
        \centering
        \includegraphics[width=0.993\textwidth,height=0.6in]{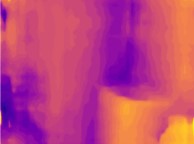}\\ 
        \includegraphics[width=0.993\textwidth,height=0.6in]{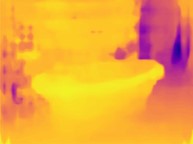}\\
        \includegraphics[width=0.993\textwidth,height=0.6in]{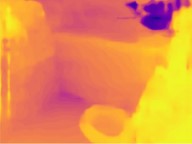}\\ 
        \includegraphics[width=0.993\textwidth,height=0.6in]{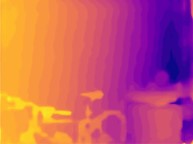}\\
        \includegraphics[width=0.993\textwidth,height=0.6in]{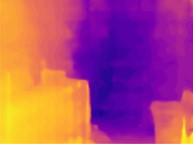}\\  
    \end{minipage}%
}%
\subfigure[BTS]{
    \begin{minipage}{0.19\linewidth}
        \centering
        \includegraphics[width=0.993\textwidth,height=0.6in]{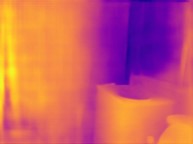}\\  
        \includegraphics[width=0.993\textwidth,height=0.6in]{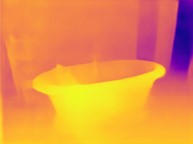}\\
        \includegraphics[width=0.993\textwidth,height=0.6in]{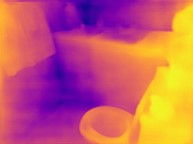}\\   
        \includegraphics[width=0.993\textwidth,height=0.6in]{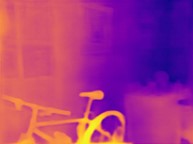}\\
        \includegraphics[width=0.993\textwidth,height=0.6in]{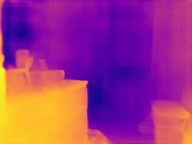}\\ 
    \end{minipage}%
}%
\subfigure[Ours]{
    \begin{minipage}{0.19\linewidth}
        \centering
        \includegraphics[width=0.993\textwidth,height=0.6in]{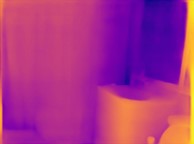}\\  
        \includegraphics[width=0.993\textwidth,height=0.6in]{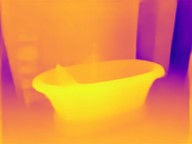}\\
        \includegraphics[width=0.993\textwidth,height=0.6in]{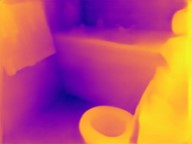}\\   
        \includegraphics[width=0.993\textwidth,height=0.6in]{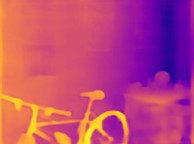}\\
        \includegraphics[width=0.993\textwidth,height=0.6in]{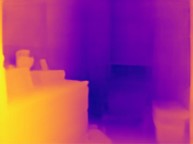}\\ 
    \end{minipage}%
}%
\centering
\caption{Qualitative examples on the NYU depth dataset.}
\label{fig:vis_nyu}
\vspace{-0.4cm}
\end{figure}

\begin{table}[t] \small
\centering
\caption{Depth Estimation: NYU dataset.}
\resizebox{1\linewidth}{!}{
\label{tab:depth_nyu}
\begin{tabular}{rcccccc}
\toprule[1.2pt]
\multirow{2.5}{*}{Method} & \multicolumn{3}{c}{Error (lower is better)} & \multicolumn{3}{c}{Accuracy (higher is better)} \\ \cmidrule(lr){2-4} \cmidrule(lr){5-7}  
 & rel & log10 & rms  & $\delta \textless 1.25$ & $\delta \textless 1.25^2$ & $\delta \textless 1.25^3$ \\ 
 \midrule
    PAD-Net~\cite{xu2018pad} & 0.214 & 0.091 & 0.792 & 0.643 & 0.902 & 0.977 \\
    Li~\etal~\cite{li2017two}& 0.152 & 0.064 & 0.611 & 0.789 & 0.955 & 0.988\\
    CLIFFNet~\cite{wang2020cliffnet} & 0.128 &	0.171 &	0.493 &	0.844 &	0.964 &	0.991\\
    Laina~\etal \cite{laina2016deeper} & 0.127 & 0.055 & 0.573 & 0.811 & 0.953 & 0.988\\
    MS-CRF~\cite{xu2017multi} & 0.121 & 0.052 & 0.586 & 0.811 & 0.954 & 0.987\\
    Lee~\etal \cite{lee2020multi} & 0.119 & 0.050 &	- &	0.870 &	0.974 & 0.993\\
    Xia~\etal \cite{xia2020generating} & 0.116 & - & 0.512 & 0.861 & 0.969 & 0.991 \\
    DORN~\cite{fu2018deep} & 0.115 & 0.051 & 0.509 & 0.828 & 0.965 & 0.992 \\
    BTS \cite{lee2019big} & 0.113 & 0.049 & 0.407 & 0.871 & 0.977 & 0.995 \\
    Yin~\etal \cite{yin2019enforcing} & 0.108 & 0.048 & 0.416 & 0.875 & 0.976 & 0.994 \\
    Huynh~\etal \cite{huynh2020guiding} & 0.108 & -	& 0.412 & 0.882 & 0.980 & \textbf{0.996}\\
    \midrule
    Baseline & 0.118 & 0.051 & 0.414 & 0.866 & 0.979 & 0.995 \\
    Ours w/ AGD & 0.111 & 0.048 & 0.393 & 0.881 & 0.979 & 0.996\\
    Ours w/ ViT & 0.109 & 0.047 & 0.388 & 0.887 & 0.981 & 0.996\\
    Ours w/ AGD+ ViT (Full)  & \textbf{0.106} & \textbf{0.045} & \textbf{0.365} & \textbf{0.900} & \textbf{0.983} & \textbf{0.996}\\
\bottomrule[1.2pt]
\end{tabular}%
}
\vspace{-0.4cm}
\end{table}

To demonstrate the competitiveness of our approach in an indoor scenario, we also evaluate the proposed method on the NYU depth dataset. The results are shown in Table~\ref{tab:depth_nyu}, compared with the the state-of-the-art methods like \cite{xu2018pad,li2017two,wang2020cliffnet,laina2016deeper,xu2017multi,lee2020multi,xia2020generating,fu2018deep,lee2019big,yin2019enforcing,huynh2020guiding}.
Similar to the experiments on KITTI, it outperforms both state-of-the-art approaches and previous methods based on attention mechanism~\cite{xu2017multi,xu2018pad,huynh2020guiding}.
Our method successfully improves~$\delta \textless 1.25$ from 0.882 (Huynh~\etal~\cite{huynh2020guiding}) to 0.900 while root mean squared error significantly drops to 0.365.
Both Table~\ref{tab:depth_kitti} and ~\ref{tab:depth_nyu} also show that our AGD can merge more low-level information and can make the network learn a more efficient deep representation.
Moreover, Figure~\ref{fig:vis_kitti} shows a qualitative comparison of our method with DORN~\cite{fu2018deep}. The red box marks the significant improvement parts. Results indicate that our method generates more precise boundaries for distant stuff like vehicles and traffic signs and near stuff like humans.
Figure~\ref{fig:vis_nyu} shows a similar comparison done on the NYU dataset. Owing to applying the Transformer, the corners of the room are more distinguishable. This can support our standpoint that adapting the linear Transformer makes the CNN backbone network enhance the ability to capture long-range dependencies. 

\begin{figure}[t] \small
\centering
\includegraphics[width=1\linewidth]{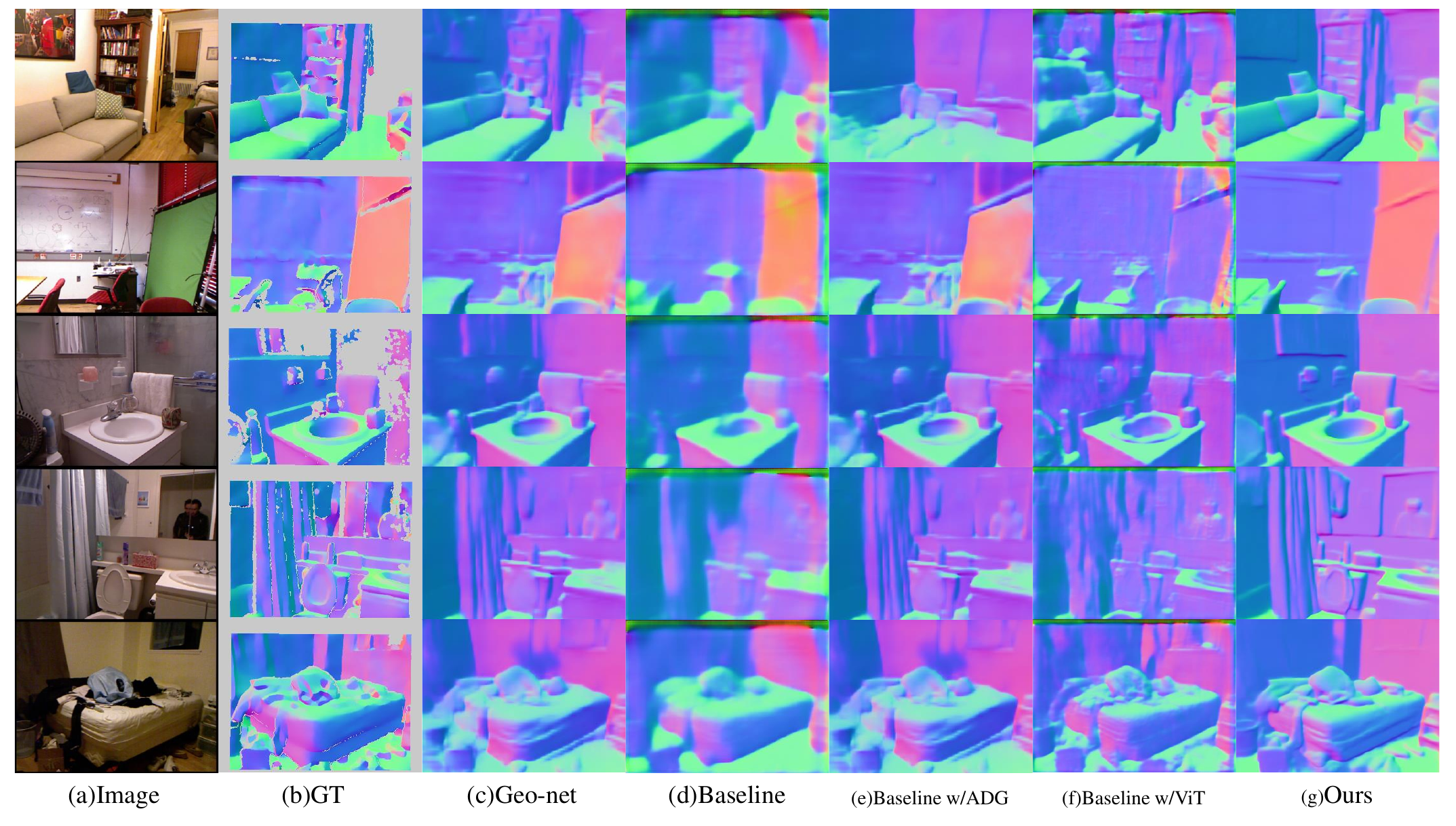}
\caption{ Qualitative examples on the NYU surface normal dataset.}
\label{fig:vis_nyu_sn}
\vspace{-0.4cm}
\end{figure}

\begin{table}[t] \small
\caption{Surface Normal Estimation: NYU dataset.}
\label{tab:sn_nyu}
\centering
\resizebox{1\linewidth}{!}{%
\begin{tabular}{r|c|c|cc}
\toprule[1.2pt]
Method & Training Data & Testing Data & median $\downarrow$ & $11.25^{\circ} \uparrow$ \\
\midrule
Li~\etal \cite{li2015depth} & \multirow{7}{*}{NYU} & \multirow{11}{*}{NYU} & 27.8 & 19.6 \\
Chen~\etal \cite{chen2017surface} &  &  & 15.8 & 39.2 \\
Eigen~\etal \cite{eigen2015predicting} &  &  & 13.2 & 44.4 \\
SURGE \cite{wang2016surge} &  &  & 12.2 & 47.3 \\
Bansal~\etal \cite{bansal2016marr} &  &  & 12.0 & 47.9 \\
GeoNet~\cite{qi2018geonet} &  &  & 12.5 & 46.0 \\
TransDepth (Ours) &  &  & \textbf{11.8} & \textbf{48.2} \\\cmidrule{1-2} \cmidrule{4-5} 
FrameNet~\cite{huang2019framenet} & \multirow{4}{*}{ScanNet} &  & 11.0 & 50.7 \\
VPLNet~\cite{wang2020vplnet} &  &  & 9.8 & 54.3 \\
Do~\etal~\cite{do2020surface} &  &  & 8.1 & 59.8 \\
TransDepth (Ours) &  &  & \textbf{7.8} & \textbf{61.7}\\
\bottomrule[1.2pt]
\end{tabular}%
}
\vspace{-0.4cm}
\end{table}

\subsection{Results on Surface Normal Estimation}
To prove our method universality, we also conduct experiments on surface normal prediction, which is regarded as a related task to depth prediction.
We compare the proposed TransDepth with several state-of-the-art methods on surface normal, including GeoNet~\cite{qi2018geonet}, VPLNet~\cite{wang2020vplnet}, FrameNet~\cite{huang2019framenet}, and Do~\etal~\cite{do2020surface}.
For a fair comparison, we report our result in two different training conditions. Because of limited space, only median angle and $11.25^{\circ}$ are compared in Table~\ref{tab:sn_nyu} while a detailed comparison is shown in the supplementary. Our method outperforms the state-of-the-art on the median angle and $11.25^{\circ}$. Though Do~\etal reduces the median angle error much, their method needs to get extra gravity labels with two-step pre-training. Our method covers these drawbacks. The qualitative results are shown in Figure~\ref{fig:vis_nyu_sn}. Unsurprisingly, the boundaries of stuff become more precise when AGD and ViT are jointly using.

\subsection{Ablation Study}
\begin{figure}[!t] \small
\centering
\subfigure{
    \begin{minipage}{0.24\linewidth}
        \centering
        \includegraphics[width=0.993\textwidth,height=0.6in]{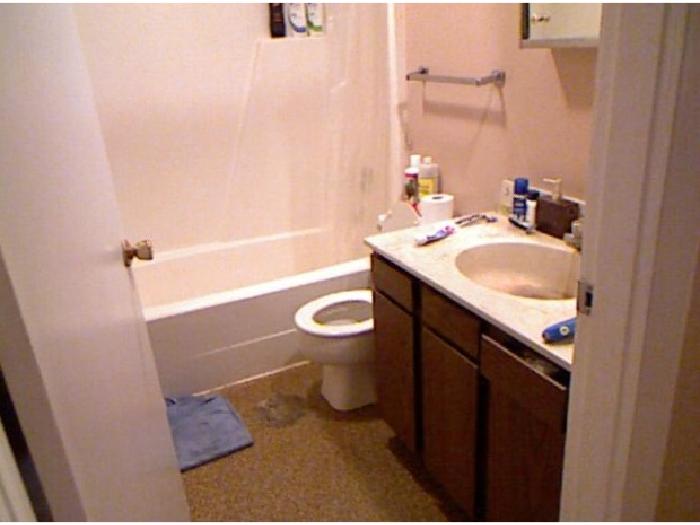}\\  
        \includegraphics[width=0.993\textwidth,height=0.6in]{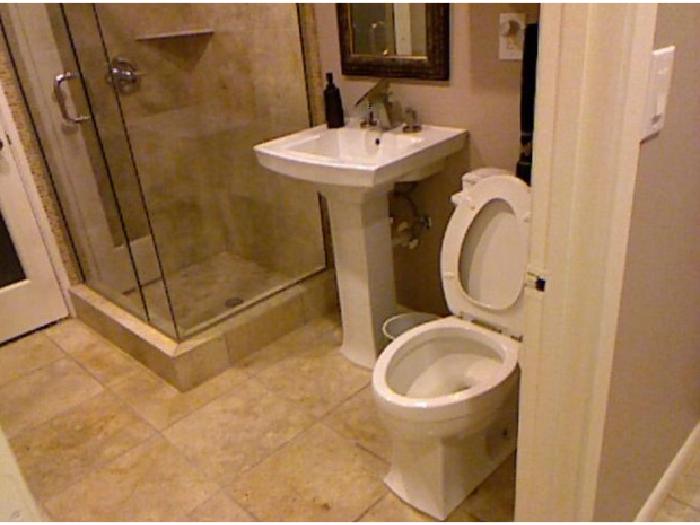}\\
        \includegraphics[width=0.993\textwidth,height=0.6in]{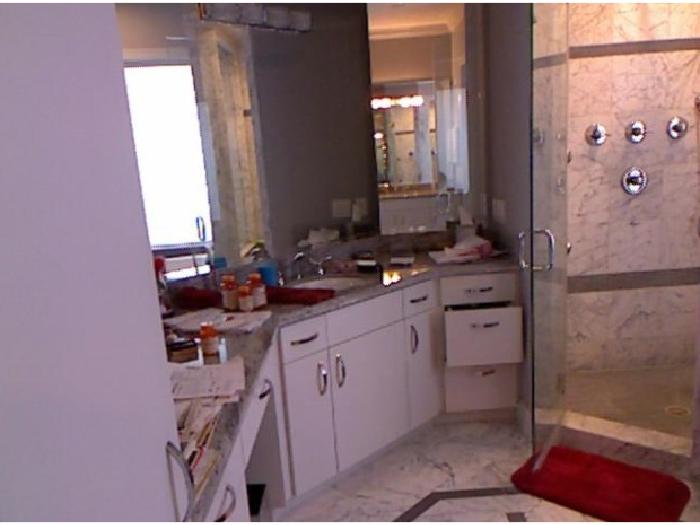}\\
        \includegraphics[width=0.993\textwidth,height=0.6in]{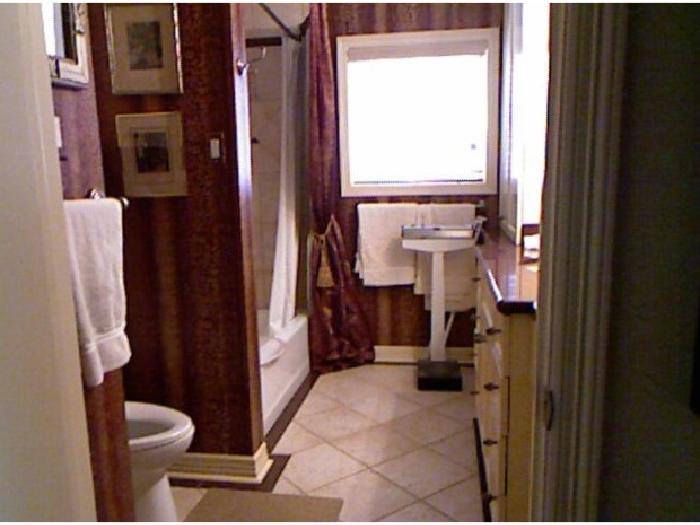}\\
        \includegraphics[width=0.993\textwidth,height=0.6in]{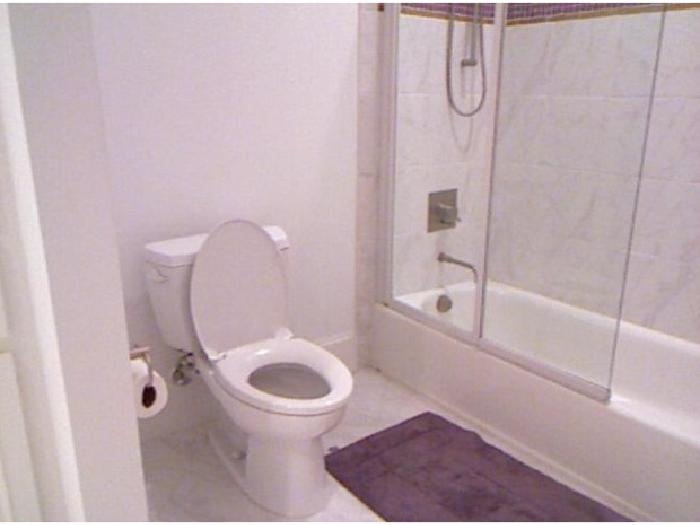}\\
    \end{minipage}%
}%
\subfigure{
    \begin{minipage}{0.24\linewidth}
        \centering
        \includegraphics[width=0.993\textwidth,height=0.6in]{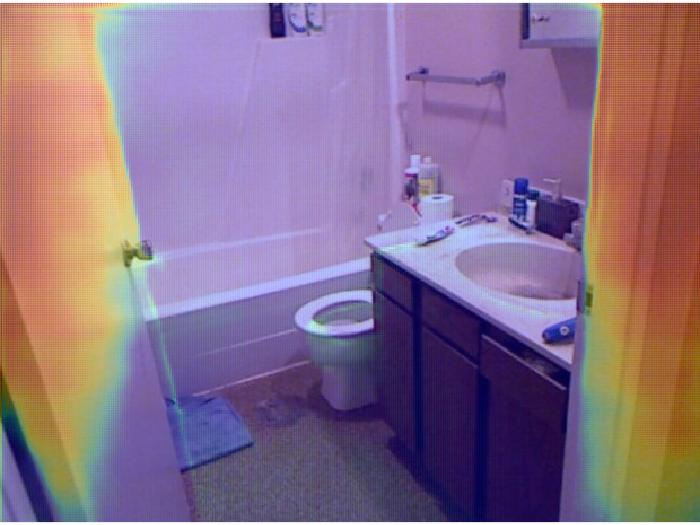}\\  
        \includegraphics[width=0.993\textwidth,height=0.6in]{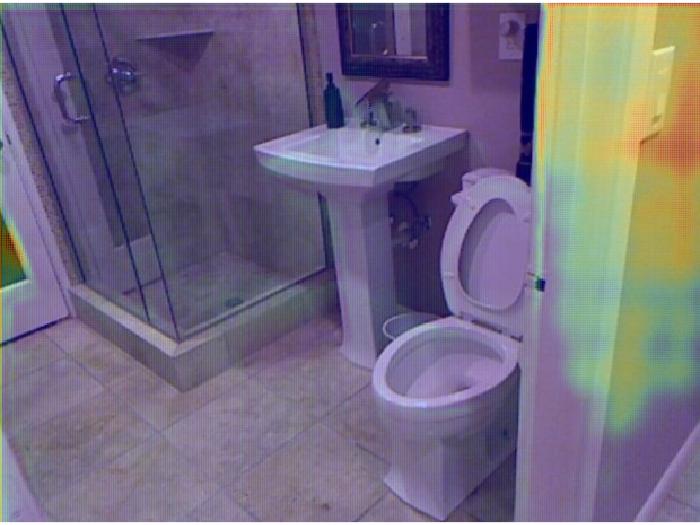}\\
        \includegraphics[width=0.993\textwidth,height=0.6in]{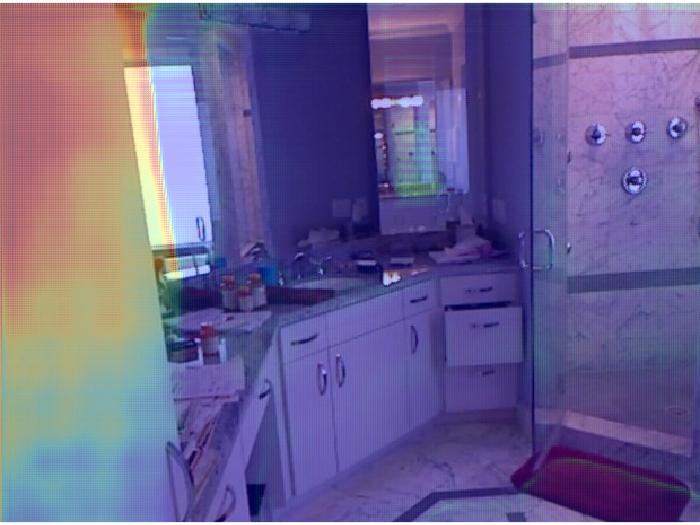}\\
        \includegraphics[width=0.993\textwidth,height=0.6in]{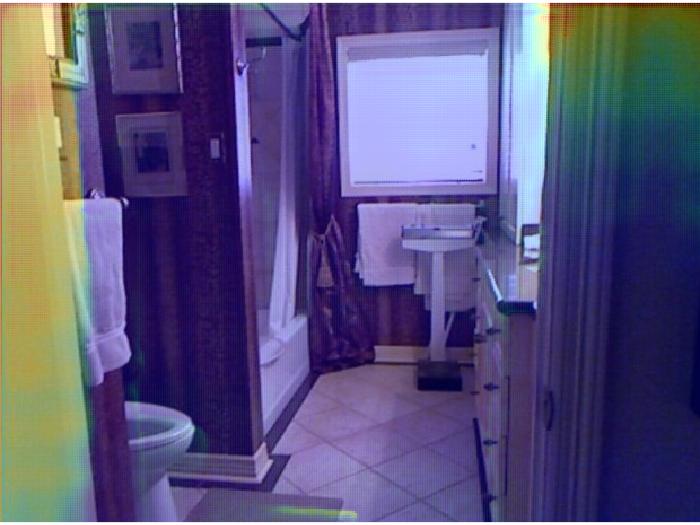}\\
        \includegraphics[width=0.993\textwidth,height=0.6in]{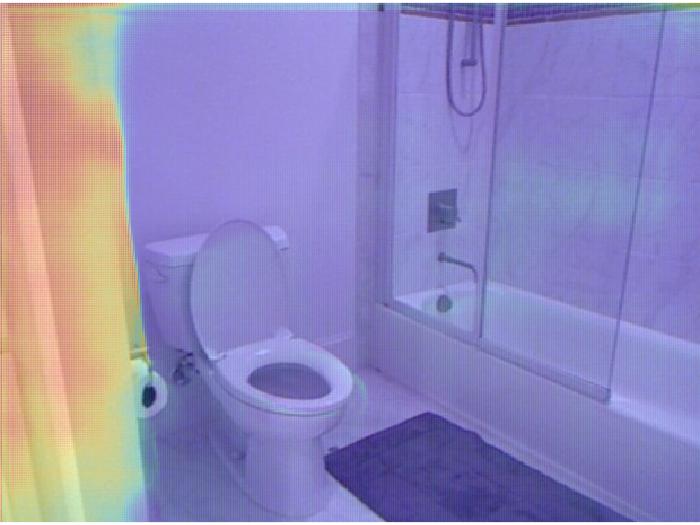}\\
    \end{minipage}%
}%
\subfigure{
    \begin{minipage}{0.24\linewidth}
        \centering
        \includegraphics[width=0.993\textwidth,height=0.6in]{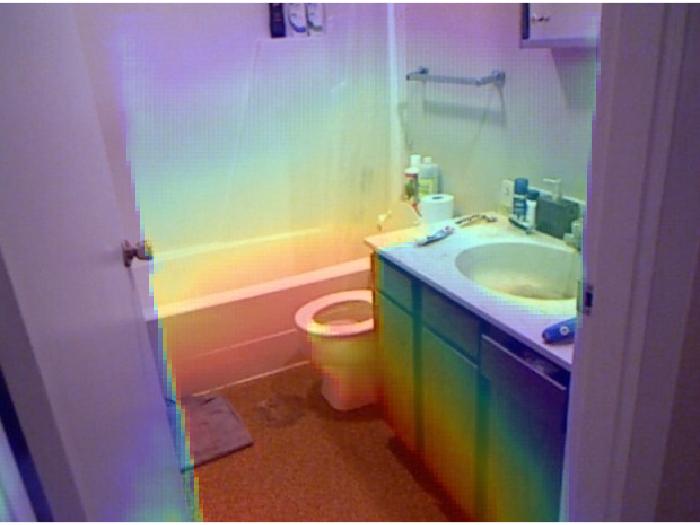}\\  
        \includegraphics[width=0.993\textwidth,height=0.6in]{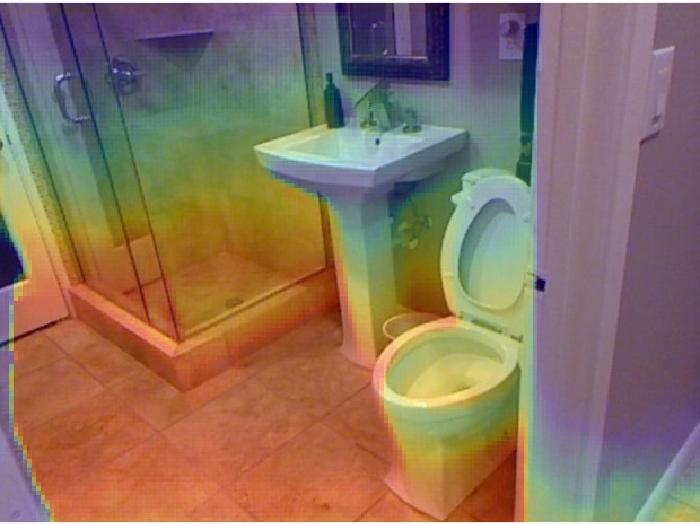}\\
        \includegraphics[width=0.993\textwidth,height=0.6in]{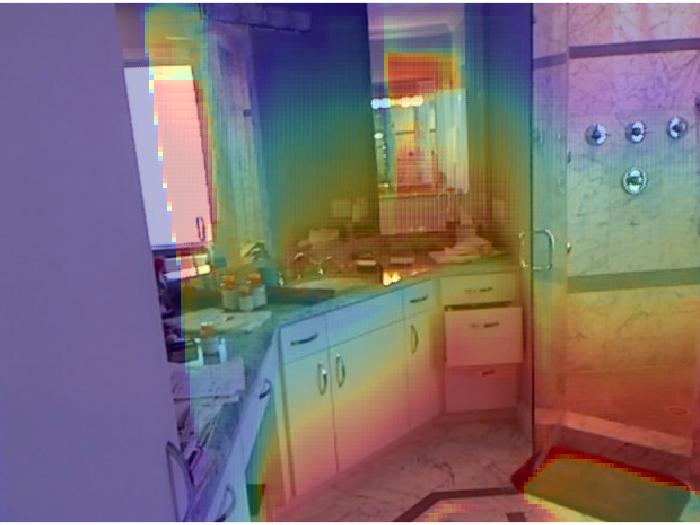}\\
        \includegraphics[width=0.993\textwidth,height=0.6in]{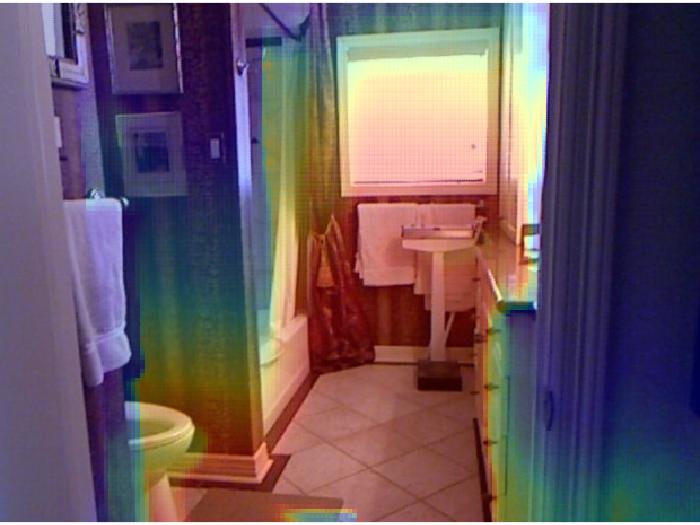}\\
        \includegraphics[width=0.993\textwidth,height=0.6in]{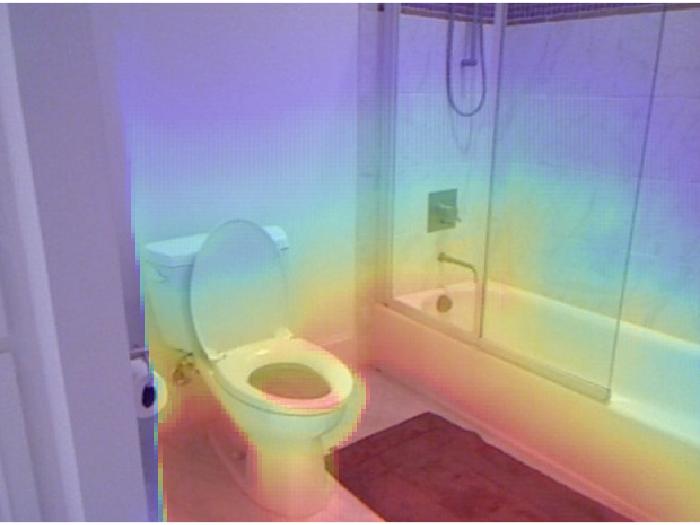}\\
    \end{minipage}%
}%
\subfigure{
    \begin{minipage}{0.24\linewidth}
        \centering
        \includegraphics[width=0.993\textwidth,height=0.6in]{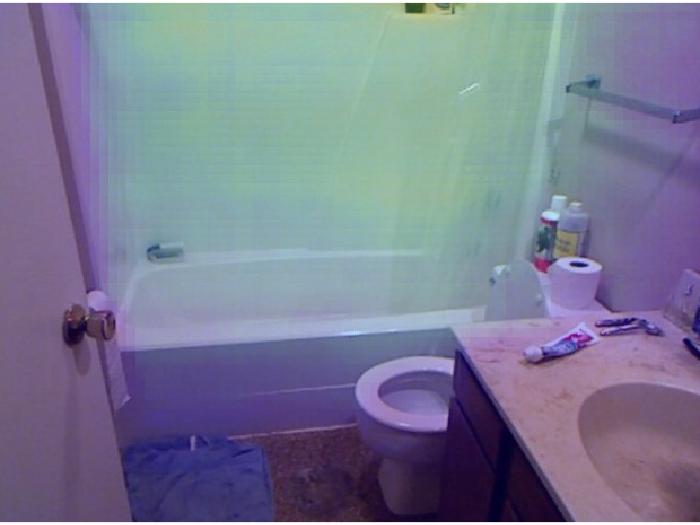}\\  
        \includegraphics[width=0.993\textwidth,height=0.6in]{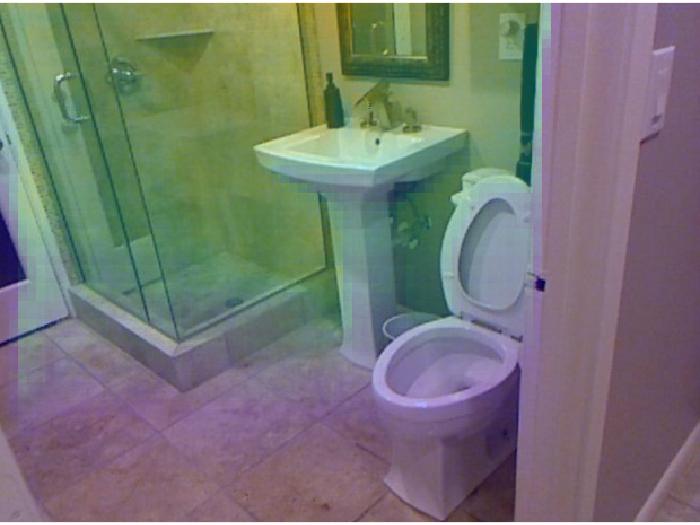}\\
        \includegraphics[width=0.993\textwidth,height=0.6in]{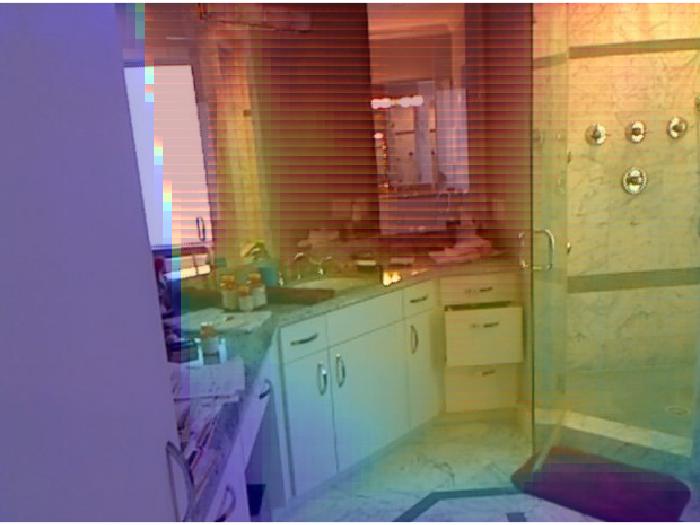}\\
        \includegraphics[width=0.993\textwidth,height=0.6in]{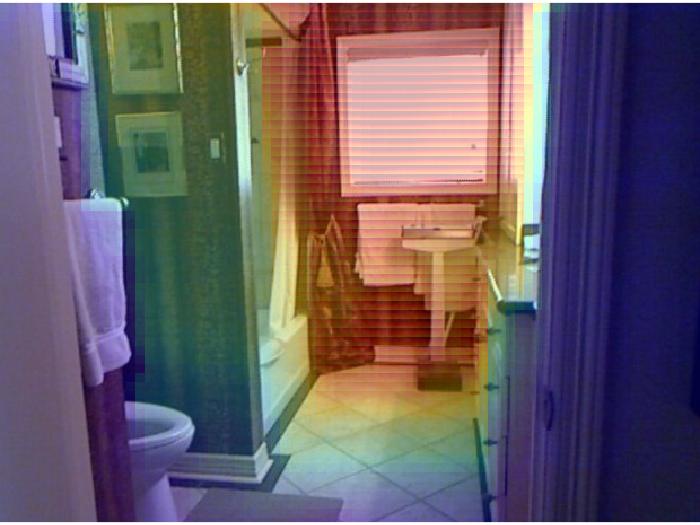}\\
        \includegraphics[width=0.993\textwidth,height=0.6in]{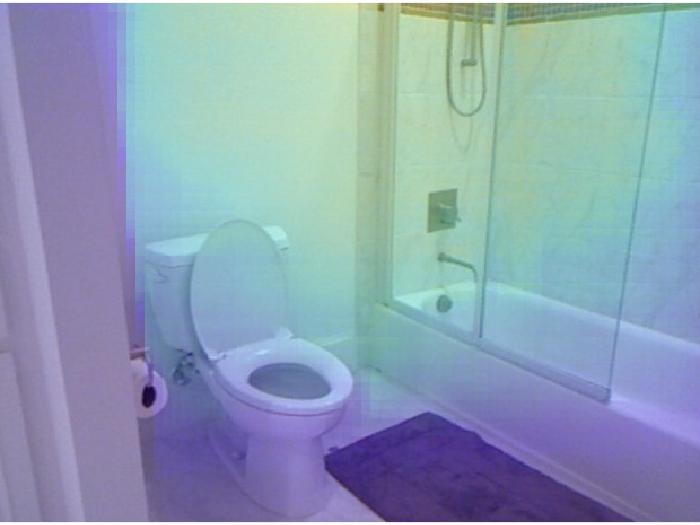}\\
    \end{minipage}%
}%
\centering
\caption{Qualitative attention examples of monocular depth prediction on the NYU dataset. The first column is the original image and the following three columns are different fusion attention.}
\label{fig:vis_att}
\vspace{-0.4cm}
\end{figure}

\noindent \textbf{Effect of Attention Gate Decoder.}
We perform an ablation study on the NYU depth dataset to further demonstrate the impact of the proposed AGD. 
In Table~\ref{tab:ab_scale}, we indicate the emitting features in the $f_e$ column while we design $f_5$, the last layer's output as the only receiving feature in all the experiments. 
We choose the ResNet-50 with the same prediction head as our baseline.
We report four different combinations with the baseline when ViT is not applied to any candidates. Interestingly, the performance does not always get better by adding more scale information. In detail, the performance increases significantly with the emitting feature increasing until the number of emitting features reaches three. Compared with the last two rows in Table~\ref{tab:ab_scale}, some metrics like rel and rms go worse when the number of emitting features further expands. This could be explained by the fact that too many scale features may lead to overfitting the receiving feature. Undoubtedly, we choose three scales of fusion in all tasks.
According to Figure~\ref{fig:vis_att}, the attention granted by different scales fusion can capture information at different range distances. This can prove that the attention gate decoder is helpful to the receive feature to capture more position information. 

\begin{table}[t] \small
\caption{Ablation study on the NYU depth dataset: performance of TransDepth for different scales fusion.}
\label{tab:ab_scale}
\resizebox{\linewidth}{!}{%
\begin{tabular}{cccccccc}
\toprule[1.2pt]
 &  & \multicolumn{3}{c}{Error (lower is better)} & \multicolumn{3}{c}{Accuracy (higher is better)} \\
 \cmidrule(lr){3-5} \cmidrule(lr){6-8}
\multirow{-2}{*}{$f_e$} & \multirow{-2}{*}{$f_r$} & rel & log10 & rms & $\delta \textless 1.25$ & $\delta \textless 1.25^2$ & $\delta \textless 1.25^3$  \\
\midrule
- & - & 0.118 & 0.051 & 0.414 & 0.866 & 0.979 & 0.995 \\
$f^5$ & $f^5 $ & 0.120 & 0.071 & 0.407 & 0.878 & 0.982 & 0.996 \\
$f^4,f^5$ & $f^5 $ & 0.108 & \textbf{0.045} & 0.366 & 0.897 & 0.982 & 0.996 \\
$f^3,f^4,f^5$ & $f^5 $ & \textbf{0.106} & \textbf{0.045} & \textbf{0.365} & \textbf{0.900} & \textbf{0.983} & \textbf{0.996} \\
$f^2,f^3,f^4,f^5$ & $f^5 $ & 0.107 & \textbf{0.045} & 0.366 & 0.899 & \textbf{0.983} & \textbf{0.996}\\
\bottomrule[1.2pt]
\end{tabular}%
}
\vspace{-0.4cm}
\end{table}

\begin{table}[t] \small
\centering
\caption{Ablation study about different backbone on the NYU depth dataset. R50 is short for ResNet50. B is short for base.}
\resizebox{1\linewidth}{!}{
\label{tab:ab_backbone}
\begin{tabular}{rcccccc}
\toprule[1.2pt]
\multirow{2.5}{*}{Backbone} & \multicolumn{3}{c}{Error (lower is better)} & \multicolumn{3}{c}{Accuracy (higher is better)} \\ \cmidrule(lr){2-4} \cmidrule(lr){5-7} 
 & rel & log10 & rms  & $\delta \textless 1.25$ & $\delta \textless 1.25^2$ & $\delta \textless 1.25^3$ \\ 
    \midrule
    ViT-B/32~\cite{dosovitskiy2020image} & 0.112 & 0.048 & 0.387 & 0.849 & 0.927 & 0.940 \\
    ViT-B/16~\cite{dosovitskiy2020image} & 0.108 & 0.046 & 0.371 & 0.885 & 0.967 & 0.979\\
    \midrule
    ResNet50~\cite{he2016deep} & 0.118 & 0.051 & 0.414 & 0.866 & 0.979 & 0.995\\
    ResNet101~\cite{he2016deep} & 0.112 & 0.048 & 0.387 & 0.848 & 0.927 & 0.939\\
    ResNet152~\cite{he2016deep} & 0.111 & 0.047 & 0.381 & 0.861 & 0.941 & 0.953\\
    \midrule
    R50+ViT-B/32  & 0.107 & 0.045 & 0.368 & 0.893 & 0.975 & 0.988\\
    R50+ViT-B/16 (Ours)  & \textbf{0.106} & \textbf{0.045} & \textbf{0.365} & \textbf{0.900} & \textbf{0.983} & \textbf{0.996}\\
\bottomrule[1.2pt]
\end{tabular}%
}
\vspace{-0.4cm}
\end{table}

\noindent \textbf{Effect of Different Backbones.}
We also compare different backbones on the NYU depth dataset on Table~\ref{tab:ab_backbone} while the attention gate decoder is not used in this experiment. The 16/32 are no longer the input path size but the shrinkage scale of the input feature. In other words, 16 means the $f^4$ is the input feature of ViT-B, while 32 represents the $f^5$ is the input feature of ViT-B.
Table~\ref{tab:ab_backbone} can be split into three-part: the top part belongs to the pure Transformer backbone; the middle part belongs to the pure ResNet backbone; the bottom belongs to the mixed backbone. 
Compared with the middle part results, the mixed backbone overpasses all of the pure ResNet backbones, leaving a significant margin for each metric. Meanwhile, according to Table~\ref{tab:ab_backbone}, the mixed backbone is better than the ResNet backbone, and it outperforms the pure Transformer encoder. 
We finally pick up the ResNet-50 with ViT-B/16 as our network's encoder for every task.

\section{Conclusions}
We propose a novel Transformer-based framework, \ie, TransDepth, for the pixel-wise prediction problems involving continuous labels. 
We are the first to propose using Transformer to solve continuous pixel-wise prediction problems to the best of our knowledge.
The proposed TransDepth leverages the inductive bias of ResNet on modeling spatial correlation and the powerful capability of Transformers on modeling global relationships. 
Moreover, a new and effective unified attention gate structure with independent channel-wise and spatial-wise attention is applied in the decoder.  This can merge more low-level information and can make the network learn a more efficient deep representation. Extensive experiments prove that the proposed TransDepth establishes new state-of-the-art results on KITTI ($0.956$ on $\delta \textless 1.25$), NYU depth ($0.900$ on $\delta \textless 1.25$), and NYU surface normal ($61.7$ on $11.25^{\circ}$) datasets.
We hope that this work can bring a new perspective on using Transformer-based architectures for computer vision tasks.

\clearpage
{\small
\bibliographystyle{ieee_fullname}
\bibliography{egbib}
}

\end{document}